\def\year{2018}\relax
\documentclass[letterpaper]{article}
\usepackage{aaai18}
\usepackage{color}
\usepackage{courier}
\usepackage{helvet}
\usepackage{times}

\usepackage{amsmath}
\usepackage{array}
\usepackage{booktabs}
\usepackage{floatrow}
\usepackage{graphicx}
\usepackage{hhline}
\usepackage{multirow}
\usepackage{subfigure}

\newcommand{\citenoun}[1]{{\citeauthor{#1} \shortcite{#1}}}
\newcommand{\cut}[1]{{}}

\newcommand*\ctoprule[1]{\addlinespace\cmidrule[\heavyrulewidth]{#1}}
\newcolumntype{M}[1]{>{\centering\arraybackslash}m{#1}}

\begin{document}
%
\title{Forecasting the presence and intensity of hostility on Instagram using linguistic and social features}
\author{Ping Liu \and Joshua Guberman\\
 Illinois Institute of Technology\\
 Chicago, IL 60616
 \And
 Libby Hemphill\\
 University of Michigan\\
 Ann Arbor, MI  48109\\
\And
 Aron Culotta\\
 Illinois Institute of Technology\\
 Chicago, IL  60616
}
 
\maketitle
\begin{abstract}
\begin{quote}
Online antisocial behavior, such as cyberbullying, harassment, and trolling, is a widespread problem that threatens free discussion and has negative physical and mental health consequences for victims and communities. While prior work has proposed automated methods to identify hostile comments in online discussions, these methods work {\it retrospectively} on comments that have already been posted, making it difficult to intervene before an interaction escalates. In this paper we instead consider the problem of {\it forecasting} future hostilities in online discussions, which we decompose into two tasks: (1) given an initial sequence of non-hostile comments in a discussion, predict whether some future comment will contain hostility; and (2) given the first hostile comment in a discussion, predict whether this will lead to an escalation of hostility in subsequent comments. Thus, we aim to forecast both the presence and intensity of hostile comments based on linguistic and social features from earlier comments. To evaluate our approach, we introduce a corpus of over 30K annotated Instagram comments from over 1,100 posts. Our  approach is able to predict the appearance of a hostile comment on an Instagram post ten or more hours in the future with an AUC of .82 (task 1), and can furthermore distinguish between high and low levels of future hostility with an AUC of .91 (task 2).
\end{quote}
\end{abstract}


\section{Introduction}
\label{sec.introduction}
Harassment, and the aggressive and antisocial content it entails, is a persistent problem for social media --- 40\% of users have experienced it~\cite{duggan2014online}. Existing approaches for addressing toxic content, including automated mechanisms, crowdsourced moderation, and user controls, have so far been ineffective at reducing hostility. Addressing harassment and other hostile online behaviors is challenging in part because people~\cite{Guberman2017-sz}, policies~\cite{Pater2016-bv}, and laws~\cite{Nocentini2010-wt} disagree about what constitutes unacceptable behavior, it takes so many forms, occurs in so many contexts, and is started by many different kinds of users~\cite{Phillips2017-ep,Phillips2015-lr,Cheng2017Anyone}.

Instead of trying to model all unacceptable content, we focus on a single class of content --- {\it hostile comments} --- that are common to a number of types of unwanted behavior including harassment and aggression. We define a hostile comment as one containing harassing, threatening, or offensive language directed toward a specific individual or group. 

Automated detection using bots or machine learning to identify and possibly deter online harassment shows promise~\cite{Sood2012-jt,Geiger2016-pi,Reynolds2011-xt}, and would greatly reduce the burden on human moderators (both lay users and moderators employed by social media platforms). However, a key pitfall of existing automated methods is that they work {\it retrospectively} on comments that have already been posted, making it difficult to intervene before an interaction escalates. In other words, users and moderators must first be exposed to the content before it can be removed. Currently, no existing tools incorporate the affordances of predictive models to \textit{prevent} harassment rather than \textit{react} to it. 
 
In this paper, we propose a method to forecast the arrival of hostile comments on Instagram posts. In order to support different intervention strategies, as well as to assess the difficulty of variants of this problem, we consider two forecasting tasks: (1) \textbf{hostility presence forecasting:} given the initial sequence of  {\it non-hostile} comments in a post, predict whether some future comment will be hostile; (2) \textbf{hostility intensity forecasting:} given the first {\it hostile} comment in a post, predict whether the post will receive more than $N$ hostile comments in the future. Solutions to the first task would support more aggressive interventions that attempt to eliminate all hostile comments from the system, while solutions to the second task would focus interventions on the most extreme cases.

We evaluate our approach on a newly constructed dataset of over 30k Instagram comments on over 1,100 posts. We identify a number of features computed over the current conversation as well as previous conversations that together provide a strong signal to forecast future hostilities in online communications. Our approach is able to predict the appearance of a hostile comment on a post ten or more hours in the future with an AUC of 0.82 (task 1), and can furthermore distinguish between high and low levels of hostility intensity with an AUC of 0.91 (task 2).


\section{Related Work}
\label{sec.related}

In this section, we review prior methods for detecting hostile content online, as well as sociological literature motivating some of the predictive features used in this study.

\subsection{Detecting hostile and antisocial content}

The three primary approaches to detecting hostile content are (1) crowd-sourced reporting, (2) moderators, (3) automated detection algorithms.
Most websites, blogs, and social media platforms crowdsource reporting and rely on their users to report or flag content as being abusive or in violation of the platform's terms and conditions. Many systems also include blocking controls that allow users to prevent others from viewing or commenting on their content. Human commercial content moderators may then attend to the flagged posts \cite{Crawford2014What}. Crowdsourced bot-based Twitter blocklists ~\cite{Geiger2016-pi} combine automation, crowdsourcing, and user controls to help users moderate their feeds. 

The flagging or reporting approach to responding to harassment exposes not only users to potentially abusive content but also the moderators who respond to it \cite{Roberts2016Commercial}. Also, like any collaborative filtering mechanism, crowdsourcing harassment detection runs the risk of “shilling attacks” ~\cite{Su2009-mx} where people give positive recommendations to their own materials and negative recommendations to others'. These reporting mechanisms also rely on free labor from members of marginalized communities, and often result in the reporters being targeted for additional harassment~\cite{Nakamura2015-mg}.

Automated algorithms may also be used to augment or replace crowdsourced moderation. In theory, automated classification of posted content would permit social media platforms to respond to greater quantities of reports in less time, thus minimizing users' exposure. To this end, effective classifiers would allow platforms to keep abreast of toxic content.   Automated mechanisms are the least common approach to addressing hostile content currently, but Instagram recently introduced tools to fight spam and encourage kinder commenting ~\cite{Thompson2017-gx}.

Prior work in this area has focused on the classification of existing, rather than future, content \cite{Yin2009Detection}. A number of classification systems have been proposed to identify toxic content, using standard text features such as TF-IDF, lexicons, and sentiment, as well as social-network features, such as number of friends/followers, and demographic attributes ~\cite{Yin2009Detection,Dinakar2011-jr,Reynolds2011-xt,Sood2012-jt,dadvar2012improved,bellmore2015five,Geiger2016-pi,al2016cybercrime}.

In this paper, we consider the more ambitious problem of forecasting future hostile content, which has the potential to allow stakeholders to stay \textit{ahead} of some toxic content. Doing so would allow alternative intervention strategies, such as issuing alerts to indicate an immediate threat and to trigger both social and technical responses to potential hostilities. Task 1's results are especially useful here where we are able to forecast whether non-hostile comments will be followed by hostile comments. Task 2's results can additionally help prioritize interventions to posts that are likely to see a sharp rise in hostile comments.

Few attempts at forecasting have been made so far, but some researchers have developed features that may be useful for this task. For instance, \citenoun{Wang2016Piece} classify \textit{disputes} using lexical, topic, discussion, and sentiment features. One of their features, \textit{sentiment transition}, estimates the probability of sentiment changing from one sentence to the next. The feature is used to augment the classifier's overall labeling capabilities. That is, rather than attempting to use sentiment transition probability to predict whether there will be a dispute within the next \textit{N} comments, the feature is used to bolster the classification of the sentiment within existing content.

Recently, \citenoun{Cheng2017Anyone} examined whether features of the context of a conversation and/or users involved could predict whether a future post will be flagged for removal by a moderator in a news discussion board. They found that an individual's mood and contextual features of a given discussion (e.g., how recently others had posted flag-worthy comments, how much time had passed since their own last comment) are related to an individual's likelihood of writing a flagged comment. Relating to discussion context, they found that an initial flagged comment predicts future flagged comments, and that the likelihood of a new discussant posting a trolling comment increases with the number of flag-worthy comments already present within a thread. While similarly motivated, there are several important distinctions between this past work and the  present study: first, we forecast hostility, rather than flagged posts; second, the best model in \citenoun{Cheng2017Anyone} assumes knowledge of the username of the author of the future flagged post, which, while useful for understanding the nature of trolling, is not a feature that would be observable by a real forecasting system. (It requires knowing who will comment on a post in the future.)

\subsection{Sociological indicators of conflict}

The literature on conflict and escalation reveal that certain patterns of interaction are more likely to lead to hostile communication, and we therefore include features that capture patterns in the comment thread. For instance, a ``conflict spiral"~\cite{Rubin1994-mi} occurs when parties mirror each other's aggressive communication tactics meaning that no matter who posts the first aggressive comment, other users will likely follow with aggressive comments of their own. Research on empathy in online communities suggests that people rarely ask for help directly~\cite{Preece1998-xk,Pfeil2007-kd}, but it is likely that help begets help in the same way that aggression begets aggression. In our Instagram scenario, this means that conversations where hostile comments are followed by positive comments may be able to recover rather than devolve into more hostility. 

Conflict spirals are less likely to occur when social bonds between parties can mediate the escalation~\cite{Rubin1994-mi}. Social similarity also mediates conflict --- in part because it is easier for us to empathize with people who are similar~\cite{Krebs1975-yh}. Relatively homogeneous groups in which people see themselves as ``in-group members" are less likely to experience hostility as well~\cite{Dovidio2013Sage}. Together, social similarity and in-group preferences mean that posts with homogeneous commenters should be less likely to receive many hostile comments. Our conversation trend and user similarity features (c.f., \S\ref{sec.forecasting}) enable us to examine the predictive utility patterns of hostility and social similarity.

Prior behavior is a useful indicator of future behavior. For instance, as mentioned above, Cheng and colleagues found that a user's previous posting history was a reasonable predictor of their future posts but that the context in which they post were better predictors~\cite{Cheng2017Anyone}. Similarly, \citenoun{Chen2012-qc}, classify both users and content, recognizing that features of both are potentially useful for reducing antisocial behaviors. We include measures of both authors' and commenters' previous posts and use different measures of time and comment thread patterns.

Many toxic remarks contain various types of profanity~\cite{Martens2015-bl,Wang2014-ir} or hate speech~\cite{Davidson2017-xm}; we use existing lexicons to generate features for both. However, we recognize that what constitutes hate speech and profanity depends on the context and geography of use~\cite{Sood2012-cn,Davidson2017-xm}.

In summary, the vast majority of prior work has focused on identifying hostile messages after they have been posted; here, we focus on forecasting the presence and intensity of future hostile comments. The primary contributions of this paper, then, are as follows:
\begin{itemize}
\item The creation of a new corpus of over 30K Instagram comments annotated by the presence of hostility (\S\ref{sec.data}).
\item A temporal clustering analysis to identify distinct patterns of hostile conversations (\S\ref{sec.cluster}).
\item The formulation of two new tasks and machine learning solutions to forecast the presence and intensity of hostile comments on Instagram posts (\S\ref{sec.forecasting}).
\item A thorough empirical analysis to understand the strengths and weaknesses of the forecasting algorithms (\S\ref{sec.experiments}-\ref{sec.results}).
\end{itemize}

\begin{table}[t]
\begin{center}
\begin{tabular}{p{25mm}M{10mm}M{15mm}M{20mm}}  
\toprule
& \textbf{posts} & \textbf{comments} & \textbf{hostile comments}\\
\midrule
\textbf{hostile posts} & 591 & 21,608 & 4,083 \\
\textbf{non-hostile posts} & 543 & 9,379 & 0 \\
\midrule
\textbf{total} & 1,134 & 30,987 & 4,083\\
\bottomrule
\end{tabular}
\caption{Statistics of the Instagram posts. Hostile posts are those with at least one hostile comment.}\label{tab.data}
\end{center}
\end{table}

\begin{figure}[t]
  \centering
  \includegraphics[width=8.5cm]{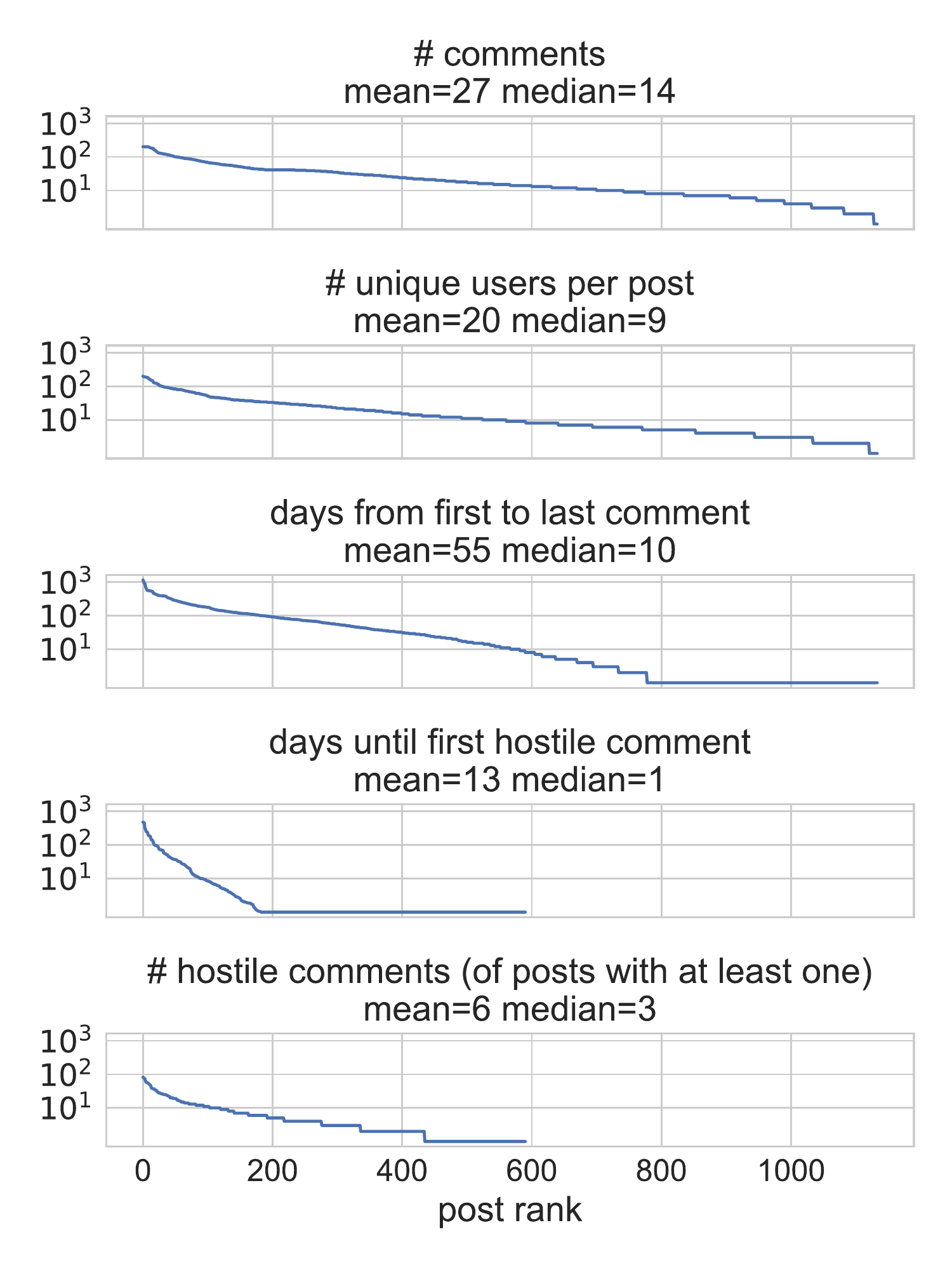}
  \caption{Rank-count plots of statistics of posts. \label{fig.ranks}}
\end{figure}

\section{Data collection and annotation}
\label{sec.data}

Given the prevalence of online harassment among teenagers~\cite{duggan2014online,Duggan2017ei}, as well as the vulnerability of this population, we decided to target our data collection to this group. According to local schools, parent groups, and recent surveys~\cite{Sood2012new}, Instagram is one of the most popular social media sites among teenagers. Instagram's predominantly public content and associated comments make it possible to collect and annotate the content necessary for our experiments.\footnote{All data and research protocols in this paper were approved by the relevant Institutional Review Board.}

We refer to a user-provided image and its associated comments as a {\it post}. Each post is authored by one user, and each post may have hundreds of comments from other users. Each comment carries a username and timestamp. To collect a sample of posts from this population, we first seeded a web crawler with the usernames of users who have public accounts, and who, upon manual inspection, appear to match the target age demographic, live in the United States, and post predominately in English. We then collected their most recent posts and associated comments. We then conducted snowball sampling by adding usernames discovered in each comment to a queue of additional users to crawl. This resulted in over 15M comments. 

For annotation purposes, we defined a hostile comment as one containing harassing, threatening, or offensive language directed toward a specific individual or group. We focus on this type of content as it is a common characteristic in many different types of antisocial, unwanted internet behavior and is a high priority for intervention. For training and validation, we required a sample of posts for which all comments are labeled with a binary indicator of hostility.

As with many other classification problems with high class imbalance, uniform sampling would result in a very low fraction of positive instances. To more efficiently allocate annotation resources, we first built a search interface to interactively explore the 15M comments retrieved via snowball sampling. After several rounds of searching and browsing, we identified a broad selection of keywords that frequently occur in hostile comments (e.g., certain profanities, abbreviations, and emojis). We then uniformly sampled 538 comments matching at least one of the identified keywords, and then downloaded all the comments from the parent post of each comment for annotation, resulting in over 20K comments. Additionally, because the experiments also require posts without any hostile comments, we sampled for annotation an additional 596 posts that did not contain any hostility keywords. (Due to the nature of our crawling procedure, we draw these non-hostile posts from similar user populations as the hostile posts.) Altogether, this resulted in 1,134 posts with 30,987 comments for annotation.

These 30,987 comments were annotated by Amazon Mechanical Turk workers who had passed certification tests indicating agreement with our labeling of example content. Each comment was annotated by two independent workers, with disagreements resolved by a third worker when needed. 

For the annotation tasks, workers were presented with an Instagram post (consisting of an image and, in some cases, a caption) and comments written in response to the post. We instructed workers to determine whether a comment was innocuous, hostile or offensive, or contained a physical threat. We also asked workers to mark comments that appear to be \textit{flashpoints}---points within already-hostile conversations at which hostility escalates precipitously, or at which a non-hostile conversation becomes markedly hostile. We directed workers to make judgments based on users' presumed intent to cause harm or distress to other users. Overall, workers were in agreement 87\% of the time when assessing whether a comment was hostile (Cohen's $\kappa=.5$).

Roughly half (591) of the posts contain at least one hostile comment, and the remainder (543) have no hostile comments. Across all data, 4,083 of 30,987 comments (13\%) are hostile. (See Table~\ref{tab.data}.)

Figure~\ref{fig.ranks} shows additional data statistics as rank-count plots (which rank posts in descending order of each statistic). 

All five statistics --- comments per post, hostile comments per post, time-duration of conversations on a post, time until the first hostile comment, and unique users per post --- exhibit long-tail properties. We also note that discussions on a post extend over long time periods --- the median post receives its final comment 10 days after the post was created. Similarly, while the first hostile comment often appears the day the post is created, 20\% of hostile posts do not receive a hostile comment until 5 or more days after they were created.

In summary, these data suggest that the commenting feature on Instagram is heavily used, with posts receiving many comments from many users over long periods of time. Furthermore, among posts containing at least one hostile comment, the volume and temporal frequency of hostile comments appear to vary substantially between posts, which motivates our tasks of forecasting both the presence and intensity of hostile comments.

\section{Identifying hostility patterns suitable for forecasting}
\label{sec.cluster}

\begin{figure}[t]
  \centering
  \includegraphics[width=8.5cm]{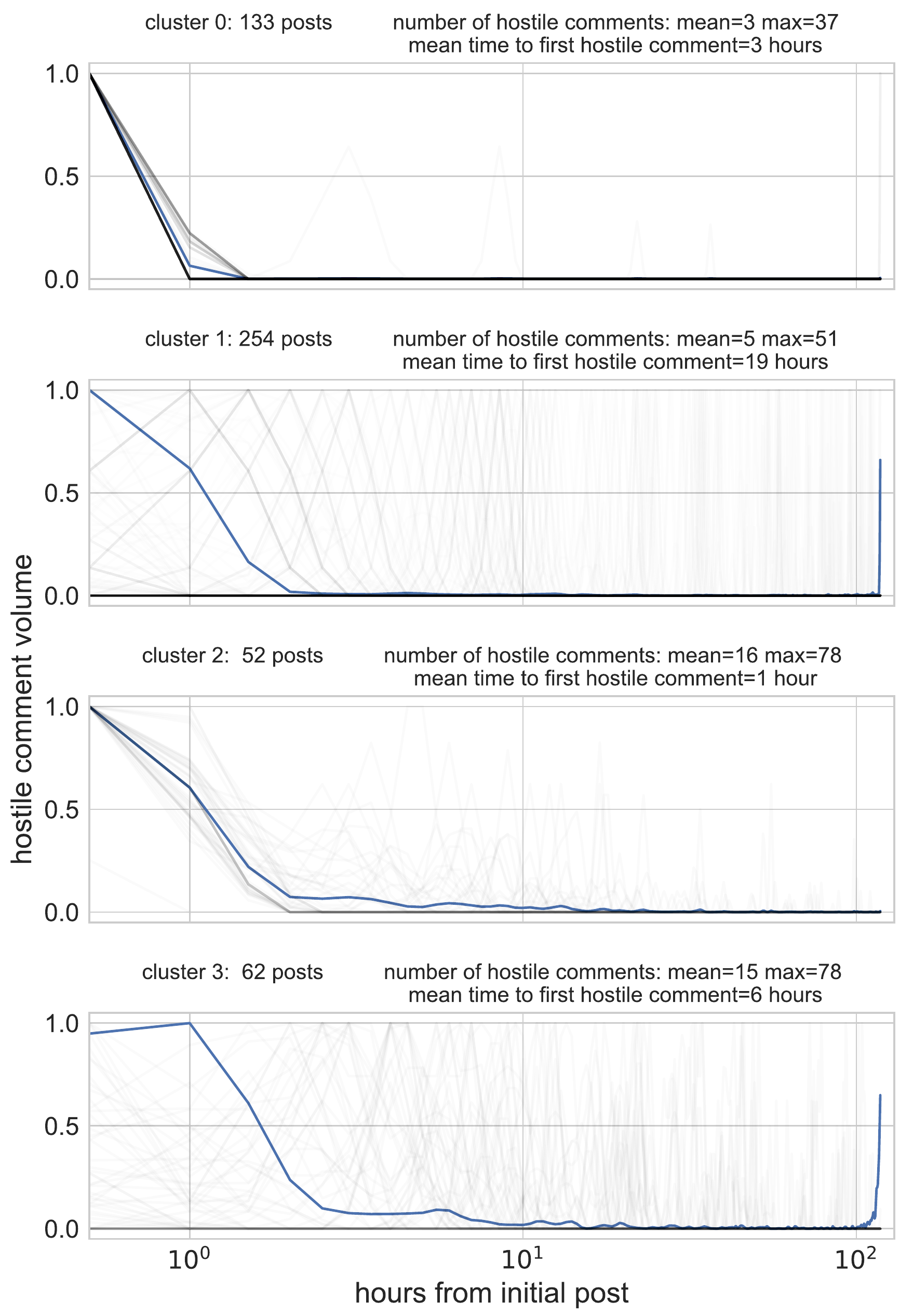}
  \caption{Volume of hostile comments over time for four discovered clusters of posts. \label{fig.clusters}}
\end{figure}

To motivate the forecasting tasks, we first explored variations in the onset and volume of hostile comments to determine which sorts of posts may be amenable to forecasting. To do so, we performed a cluster analysis to identify groups of posts that exhibit similar temporal characteristics. We first constructed a time series for each post by counting the number of hostile comments posted within each hour from the time the post was created.\footnote{ We smooth each series with a Gaussian kernel (width=5, $\sigma$=1).} We restricted this analysis to the 501 posts that have at least one hostile comment within the first 10 days. We then clustered the time series using the K-Spectral Centroid (K-SC) clustering algorithm of \citenoun{yang2011patterns}. K-SC clusters a set of time series using a distance measure that is invariant to scaling and shifting, enabling us to identify posts that have similarly shaped time series of hostile post frequency.

Figure~\ref{fig.clusters} shows four discovered clusters of posts.\footnote{Increasing the number of clusters up to 10 had similar results.} 

Each gray line corresponds to the smoothed time series for a single post, and the blue lines indicate the centroid of each cluster. The y-axis is scaled to 1 by dividing each series by its maximum value. We can see four distinct patterns. Cluster 0 has the lowest level of hostility (only three hostile comments, on average), and these hostile comments arrive almost immediately after the post is created. Cluster 1 also has a relatively small number of hostile comments, but these can arrive much later (19 hours after the post was created, on average). Clusters 2 and 3 show a similar pattern, but for posts with roughly three times the volume of hostile posts. Thus, clusters 1 and 3 both have a delay in the onset of the first hostile message, but cluster 3 results in a much higher volume of hostile messages in the remainder of the comments. 

These results provide some insight into what sorts of interventions may have an impact and what their limitations may be. For example, posts from cluster 0 appear to be a low priority for intervention --- the hostile comments are isolated incidents that occur too quickly for any preventative action. In contrast, in clusters 1 and 3 the time lag before hostile comments appear presents an opportunity to intervene before hostilities escalate --- e.g., by blocking comments, taking down a post, or other means. This motivates our first task of forecasting a future hostile comment in a post. Additionally, while the first hostile comment appears similarly quickly in clusters 0 and 2, if we could distinguish between the two after the first hostile comment appears, we could prioritize posts from cluster 2 for intervention. This motivates our second task of forecasting the total volume of hostile comments a post will eventually receive.

\section{Forecasting hostility}
\label{sec.forecasting}

In this section, we formalize our two forecasting tasks and our proposed approach. Our data consist of $m$ posts $\{p_1 \ldots p_m\}$, where post $p_i$ is an ordered list of tuples $p_i=\{(c_1, t_1, y_1) \ldots (c_n, t_n, y_n) \}$, with $(c_j, t_j, y_j)$ indicating the $j$th comment, timestamp, and hostility label, respectively. We assume $y_j=1$ indicates a hostile comment, and $y_j=0$ otherwise. Given the patterns observed above, we consider the following two forecasting tasks:

{\bf Task 1: Hostility presence forecasting.} Given an initial sequence of {\it non-hostile} comments on a post, predict whether at least one future comment will be hostile. That is, for each post the system observes $\{(c_1, t_1, y_1) \ldots (c_j, t_j, y_j)\}$ where $y_i=0$ $\forall \: i\le j$ and must predict whether there is some subsequent comment $k>j$ such that $y_k=1$.

We refer to $t_k - t_j$ as the {\it lead time}: the amount of time between the final comment observed by the system and the time of the first hostile message. From an intervention perspective, greater lead times are desirable, since they provide more opportunity to take preventative measures. However, we expect problem difficulty to increase with required lead times, which we investigate below.

{\bf Task 2: Hostility intensity forecasting.} Given all comments up to and including the first hostile comment, predict whether the total number of hostile comments on a post will be greater than or equal to a given threshold $N$. That is, the system observes $\{(c_1, t_1, y_1) \ldots (c_j, t_j, y_j)\}$, where $y_j=1$ and $y_i=0$ $\forall \: i<j$. A post containing $m$ comments has a positive label if $\sum_{k=1}^m y_k \ge N$. When $N$ is large, the system is designed to identify posts with a high intensity (volume) of hostile comments, which can then be prioritized for intervention.

Because we frame both tasks as binary classification problems, we model each task using logistic regression with L2 regularization.\footnote{Additional experiments with a Random Forest classifier produced similar results.} We next describe the features used for this classifier. 

\subsection{Features}

For both tasks, we compute features over the linguistic context of the current post and the historical context of the users participating in the post.

\begin{itemize}
\item \textbf{Unigram (U)} We tokenize each comment into unigrams, retaining emojis as separate words, and collapsing all user  @mentions to a single feature type. Multiple comments are collapsed into a single bag-of-words vector.

\item \textbf{Word2vec (w2v)} To overcome the sparsity of unigram features, we fit a word2vec model~\cite{mikolov2013distributed} on the $\sim$15M unlabeled comments collected in our initial crawl. We set the vector length to 100 dimensions, and to aggregate the vectors for each word in a comment, we concatenate the maximum and average along each word vector dimension (i.e., 200 total features).
\item \textbf{n-gram word2vec  (n-w2v)} \citenoun{bojanowski2016enriching} introduce an extension to word2vec designed to model character n-grams, rather than words, which may help when dealing with short social media messages that contain abbreviations and spelling errors. We train on the same 15M comments using this n-gram word2vec model and again use maximum and average vectors to aggregate across words in a sentence.
\item \textbf{Hatebase/ProfaneLexicon (lex)} We include features indicating whether a comment contains words present in two lexicons of offensive language: Hatebase\footnote{www.hatebase.org} and  Luis von Ahn's Offensive/Profane Word List\footnote{www.cs.cmu.edu/$\sim$biglou/resources/}. Hatebase has 582 hate words in English, and categorizes each hate word into six different categories: class, disability, ethnicity, gender, nationality and religion. For each category, we have a binary feature that is one if the sentence has at least one word in Hatebase; we add an additional feature for the total count across all categories. The Offensive/Profane Word List provides 1,300+ English terms that may be considered offensive; we include both binary and count features for this lexicon as well.
\item \textbf{Final comment (final-com)} For posts with many comments, it may be necessary to place greater emphasis on the most recent comment. To do so, we create a separate feature vector using only the most recent comment, using unigram, n-gram word2vec, and lexicon features. E.g., one such feature may indicate that a word from the Hatebase lexicon appeared in the most recent observed comment for this post.
\item \textbf{Previous comments (prev-com)} Prior work suggests that a user's recent posts may indicate their propensity to post offensive comments in the near future~\cite{Cheng2017Anyone}. To capture this notion, we first identify the usernames of the authors of each comment the system observes in the target post. We then identify the most recent comment each user has made in a post other than the target post. We then concatenate all these comments together, building unigram, n-gram word2vec and lexicon features. These features serve to summarize what the commenters in the current post have recently said in prior posts.
\item \textbf{Previous post (prev-post)} Conversely, certain users' posts may be more likely to receive hostile comments than others, perhaps because of their content, recent adversarial interactions with other users, or other attributes of the user. To capture this, we collect the most recent post made by the author of the target post and concatenate all the comments in that post, again using unigram, n-gram word2vec and lexicon features. These features serve to summarize the recent conversations involving the author of the target post. 
\item \textbf{Trend Feature (trend)} While the linguistic features summarize the context of the current conversation, they ignore the order of the comments. The goal of this feature type is to detect an increase over time in the hostilities expressed in the current post to identify trends that precede hostile comments. To do so, we first train a separate logistic regression classifier to determine if a comment expresses hostility or not, using the same unigram, n-gram word2vec, and lexicon features from above (it obtains cross-validation AUC of $\sim$.8). We then apply this classifier to each comment observed in the target post, recording the predicted posterior probability that each comment is hostile.\footnote{This classifier is trained with double cross-validation so that it never predicts the label of a comment that was in its training set.}  We then compute several features over this vector of posterior probabilities:
(1) the number and fraction of comments that have positive probability larger than some threshold (0.3 in the experiments below); (2) the maximum slope within two adjacent probabilities of positive class (to detect sudden shifts in the conversation); (3) the difference between the maximum and minimum probability across all comments.
\item \textbf{User activity (user)} Finally, inspired by evidence that group heterogeneity can increase likelihood of hostility \cite{Dovidio2013Sage},
we compute two additional features based on the number of users who have commented on the target post: (1) the number of unique users divided by the number of comments (to quantify the heterogeneity of comment authors); and (2) the fraction of comments containing a  user @mention (to quantify the volume of directed utterances).
\end{itemize}

\section{Experiments}
\label{sec.experiments}

We conduct 10-fold cross-validation experiments to measure the forecasting accuracy for each task.\footnote{For replication code, see https://github.com/tapilab/icwsm-2018-hostility} For Task 1, the system observes the first comments of a post and predicts whether some hostile comment will appear. Rather than fixing the number of comments observed, we instead vary the lead time, defined as the time from the final comment observed on a post to the time of the first hostile comment. For example, if the first hostile comment on a post occurs at 3 p.m., a lead time of 1 hour means that the forecasting system will observe all comments on the post made before 2 p.m. We consider lead times of $\{1,3,5,8,10\}$ hours. Since the optimal classifier may vary by lead time, we train separate models for each lead time. We chose these lead time intervals based on our observations of the pace at which comments are posted and when hostile comments first appear (see Figure~\ref{fig.clusters}).

For some lead times, posts may be discarded from experiments --- e.g., if the lead time is 10 hours, but the first hostile message occurs in the first hour, then it is not possible to observe any comments in time for that post. Similarly, the lead time also affects the number of comments the system may observe on average (i.e., longer lead times typically mean fewer comments may be observed before making a prediction). We sample an equal number of positive and negative examples for this task; to avoid introducing bias between classes, positive and negative samples are matched by the number of comments (e.g., for a positive example where we observe six comments, we sample a matched negative example from the set of non-hostile posts and restrict the observation to the first six comments.)

For the second task (hostility intensity forecasting), the system observes the first comments of a post, up to and including the first hostile comment, and predicts whether there will be $\ge N$ total hostile comments on the post. Thus, $N$ serves as a threshold for instances to be considered in the positive class. From a practical perspective, it is more useful to distinguish between posts with 1 versus 10 hostile comments rather than 1 versus 2 hostile comments. We thus vary $N$ from 5-15, training separate models for each setting. For a given $N$, positive instances must have $\ge N$ hostile comments, while negative instances must have only 1 hostile comment (posts with other values are not included in this analysis). Thus, the goal is to separate ``one-off" hostile comments from those that result in an escalation of hostility.

\begin{figure}[t]
  \centering  
  \includegraphics[width=8.5cm]{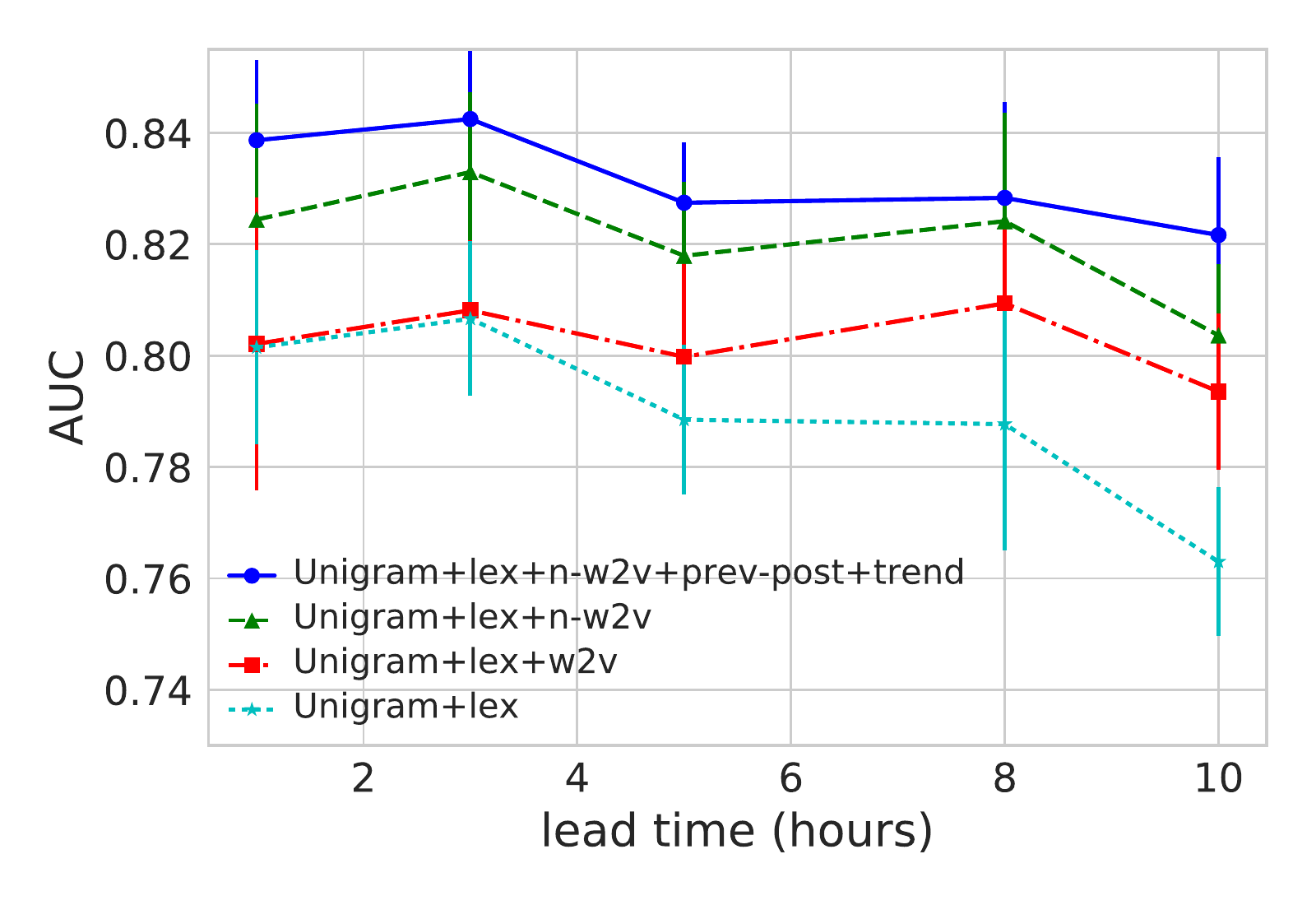}
  \caption{Hostility presence forecasting accuracy as lead time increases. \label{fig.forecast_time}}
\end{figure}

\begin{table}[t]
  \centering
  \begin{tabular}{p{25mm}M{6mm}M{6mm}M{12mm}M{12mm}}
    \ctoprule{1-5}
    \textbf{Features}  & \textbf{AUC}  & \textbf{F1} & \textbf{Precision} & \textbf{Recall} \\
    \midrule
\textbf{Unigram      } & 0.790 & 0.737 &      0.723 &   0.778 \\
\textbf{U + prev-com } & 0.707 & 0.672 &      0.660 &   0.693 \\
\textbf{U + final-com} & 0.779 & 0.729 &      0.716 &   0.757 \\
\textbf{U + trend    } & 0.790 & 0.739 &      0.725 &   0.776 \\
\textbf{U + user     } & 0.791 & 0.742 &      0.729 &   0.778 \\
\textbf{U + lex      } & 0.792 & 0.750 &      0.732 &   \textbf{0.788} \\
\textbf{U + w2v      } & 0.794 & 0.725 &      0.715 &   0.749 \\
\textbf{U + n-w2v    } & 0.810 & 0.736 &      0.725 &   0.746 \\
\textbf{U + prev-post} & 0.828 & 0.761 &      \textbf{0.756} &   0.765 \\
\midrule
\textbf{Best         } & \textbf{0.843} & \textbf{0.765} &      0.755 &   0.778 \\
   \bottomrule
  \end{tabular}
  \caption{Forecasting accuracy of Task 1 (lead time = 3 hours). The best combination uses all features except for w2v and prev-com.}
  \label{tab.forecast_results_task_1}
\end{table}

\begin{figure}[t]
  \centering  
  \includegraphics[width=8.1cm]{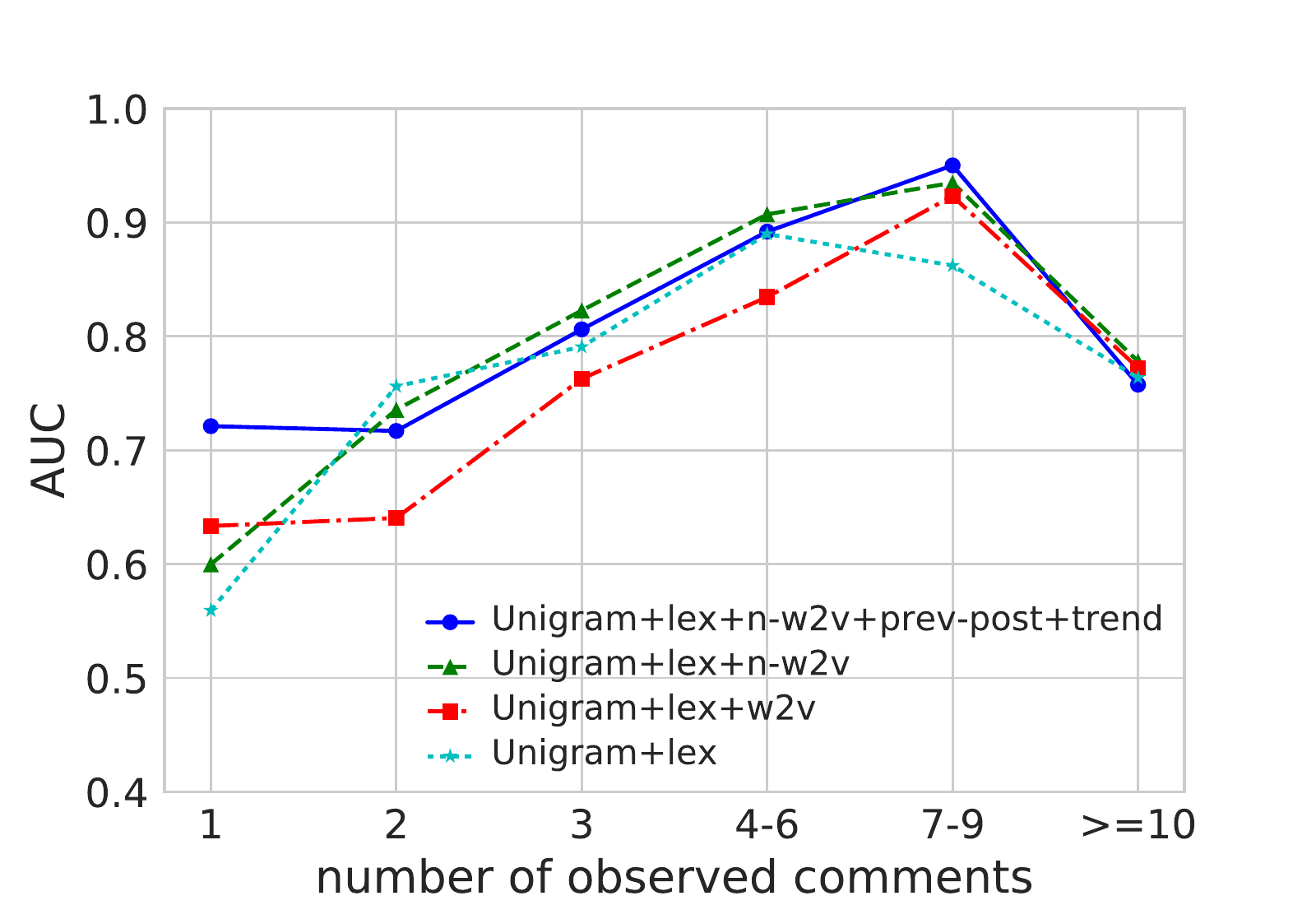}
  \caption{Hostility presence forecasting accuracy as observed comments increase (lead time=3 hours). \label{fig.forecast_comments}}
\end{figure}

Because both tasks can be viewed as ranking problems (to prioritize interventions), our primary evaluation metric is AUC (area under the ROC curve). For comparison, we also report F1, Precision, and Recall.

\section{Results}
\label{sec.results}

\subsection{Task 1: Presence forecasting}

Figure~\ref{fig.forecast_time} shows the AUC for four feature settings as the lead time increases. The best model achieves a high AUC for a lead time of one hour (0.84). There is a steady drop in accuracy as lead time increases, as expected. However, even with a 10 hour lead time, the best model only declines by 2\% from a lead time of one hour (0.84 to 0.82). This suggests that there is sufficient linguistic context early in a post to indicate whether it will eventually attract hostile comments. This figure also suggests that the n-gram word2vec features are somewhat more predictive than the traditional word2vec features across all lead times.

Table~\ref{tab.forecast_results_task_1} provides a detailed feature comparison for a lead time of 3 hours. We can see that the previous post features are the most useful addition to the unigram features, increasing AUC from .790 to .828. Thus, users who have received hostile comments in the past are more likely to receive them in the future. In contrast, the previous comment features are unhelpful --- including them actually reduces AUC from .790 to .707. Upon further inspection, we attribute this to the fact that very often (494 of 591 hostile posts) the first hostile comment on a post is from a user who has not previously commented on this post. Thus, including features from the users who have commented on the post already may tell us little about the user who will initiate hostility. Together, these results suggest that certain users are more susceptible to receiving hostile comments.

Figure~\ref{fig.forecast_comments} stratifies the results of the four models by the number of observed comments, where the lead time is 3 hours. We can see that the number of observed comments can greatly affect accuracy. The best model has an AUC of .71 when observing only one comment, but an AUC of .92 when observing 7-9 comments. In practice, one may wish to restrict forecasts to posts with a sufficient number of observed comments. Furthermore, it appears that the previous post features most improve over baselines when there are few observed comments, presumably by providing additional context that is missing when there are few comments.

Figure~\ref{fig.forecast_comments} somewhat surprisingly shows a decrease in AUC when observing 10 or more comments. By reviewing these posts, we observe several instances of what we call ``non-sequitur hostility." For example, a post may quickly receive 20 innocuous, non-controversial comments (e.g., congratulating the poster on some achievement). Subsequently, a new user writes a hostile comment that is unrelated to the preceding conversation. These instances are very difficult to predict -- since the model observes many innocuous comments, it assigns a low probability of seeing a hostile comment in the future. It is possible that accuracy may be improved by training separate classifiers after grouping by the number of observed comments.

\subsection{Task 2: Intensity forecasting}
\begin{table}[tp]
  \centering
    \begin{tabular}{p{25mm}M{6mm}M{6mm}M{12mm}M{12mm}}
    \ctoprule{1-5}
         \textbf{Features}  & \textbf{AUC}  & \textbf{F1} & \textbf{Precision} & \textbf{Recall} \\
      \midrule
\textbf{Unigram      } & 0.808 & 0.747 &      0.741 &   0.673 \\
\textbf{U + w2v      } & 0.753 & 0.696 &      0.662 &   0.673 \\
\textbf{U + prev-com } & 0.786 & 0.701 &      0.694 &   0.605 \\
\textbf{U + user     } & 0.817 & 0.761 &      0.752 &   0.695 \\
\textbf{U + n-w2v    } & 0.821 & 0.775 &      0.781 &   0.711 \\
\textbf{U + trend    } & 0.825 & 0.778 &      0.782 &   0.721 \\
\textbf{U + lex      } & 0.827 & 0.776 &      0.785 &   0.705 \\
\textbf{U + prev-post} & 0.842 & 0.782 &      \textbf{0.829} &   0.688 \\
\textbf{U + final-com} & 0.879 & 0.792 &      0.805 &   0.722 \\
\midrule
\textbf{Best         } & \textbf{0.913} & \textbf{0.805} &      0.785 &   \textbf{0.772} \\
\bottomrule
   \end{tabular}
  \caption{Forecasting accuracy of Task 2 (N=10). The best feature combination is trend/user/final-com.}
  \label{tab.forecast_results_task_2}
\end{table}

\begin{figure}[t]
  \centering  
  \includegraphics[width=8.5cm]{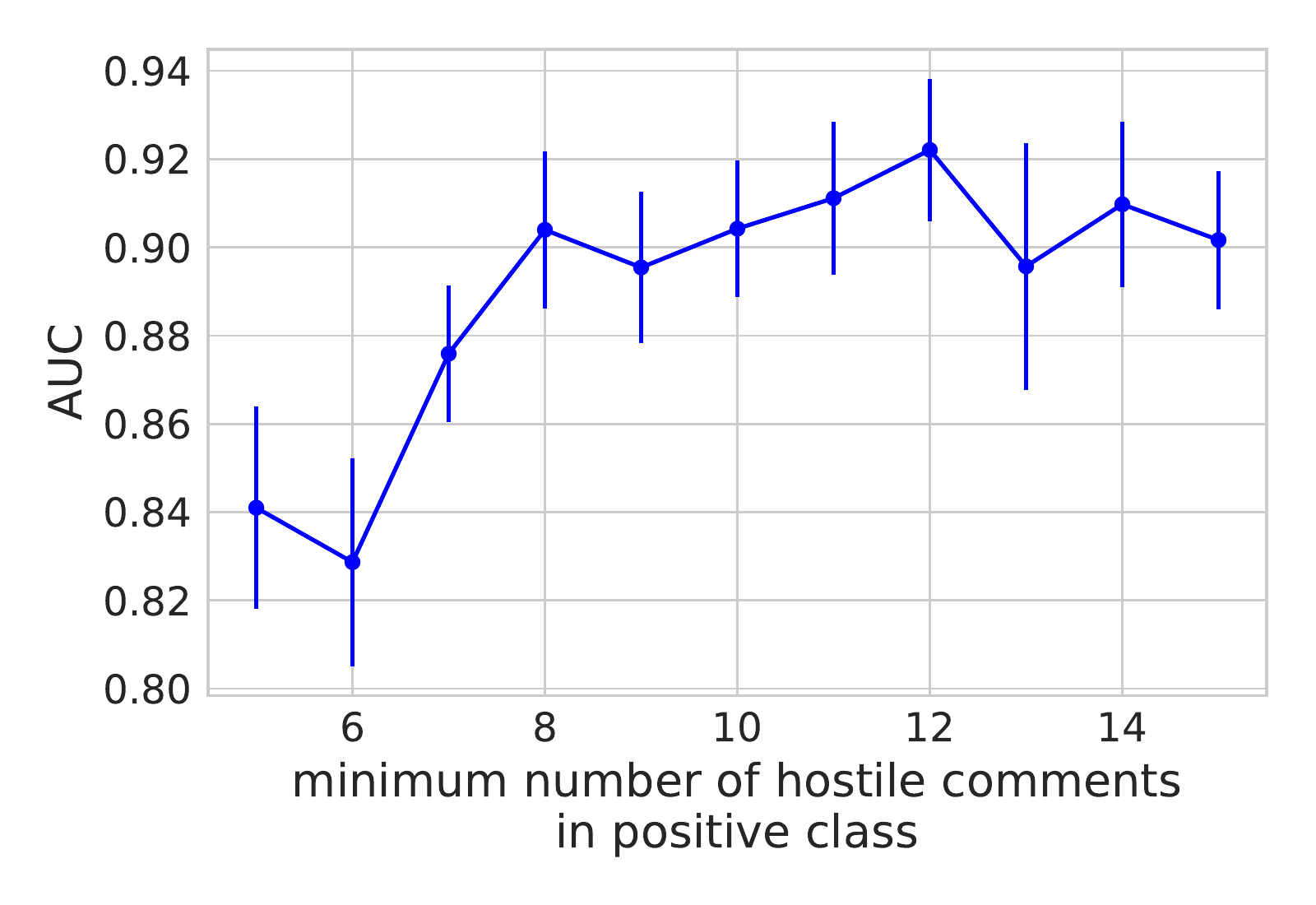}
  \caption{Hostility intensity forecasting accuracy as the positive class threshold increases (with standard errors). \label{fig.intensity_threshold}}
\end{figure}

Figure~\ref{fig.intensity_threshold} shows the AUC for the best model as the positive class threshold increases. As expected, it is easier to identify posts that will receive 10 or more hostile comments than those that will receive only 5 or more. Table~\ref{tab.forecast_results_task_2} reports results from different feature combinations using 10 as the positive class threshold. As in Task 1, the  previous post features lead to improved AUC (.842 vs .808), but the previous comment features do not (.786 vs. .808). These results suggest that it is possible to distinguish between posts that will receive an isolated hostile comment and those that will receive high volumes of hostile comments. Thus, interventions may be prioritized by the expected intensity of hostile interactions.

To better understand the which features are leading indicators of a high volume of hostile comments, we examined the top coefficients for the positive class for $N=10$. Overall, the final-comment features are the most predictive --- it seems that features of the first hostile message are predictive of the volume of hostile message in the future. Examining the top coefficients for these features, we find that if the first hostile comment contains a user mention, a second-person pronoun, or profanity, it is correlated with a high intensity post. Other top features include: (1) a term matching the Profane Lexicon was used in a comment made on an earlier post by this author; (2) a comment on the current post contains a term from the Gender category in the HateBase lexicon; (3) the abbreviation ``stfu" (``shut the f**k up"), which indicates a possible turning point in the conversation leading to an escalation in hostilities; (4) singular second- and third-person pronouns like ``you" and ``she," which indicate that the conversation is directed toward specific users rather than distant third parties.

We also examined a sample of forecasting errors. False negatives often occur when many similar innocuous messages are posted consecutively, giving no indication of any escalation. For example, in one post, the first thirty comments were short messages wishing the author a happy birthday. The first hostile comment insulted the post author, resulting in many retaliatory messages from his friends. However, given the limited information in the first comments, the system did not predict this escalation in hostility.

This example demonstrates the nuance of online hostilities --- it is rare in our data to see clear victims and perpetrators. Instead, it is more common to see an outsider write insulting messages to an individual, which in turn leads others to make insulting responses, which in turn escalates into a back-and-forth of hostile messages.

\section{Limitations}
As described in \S\ref{sec.data}, the data is not a uniform random sample of all Instagram posts. Thus, the observed frequency of hostile comments should not be interpreted as an estimate of the true fraction of hostile comments on Instagram. Instead, we focused our limited annotation resources on posts that are (1) likely to be from vulnerable populations (teenagers) and (2) likely to contain hostility. To do so, we used a broad set of keywords to find an initial set of 538 comments (and corresponding posts), which we then augmented with 596 posts that did not contain such keywords. Although the seed sample of 538 comments is influenced by the keywords chosen, by annotating all the other comments in each post we also capture many additional hostile comments that do not match the initial set of keywords. Indeed, over 85\% of the hostile comments in the final data were not retrieved by the initial keyword search. Furthermore, of the 538 posts containing a hostile keyword, only 504 were annotated as hostile; of the 596 posts that did not contain a hostile keyword, 87 had at least one comment annotated as hostile. Thus, the data contain many hostile comments that do not match the keywords, as well as many non-hostile comments that do match the keywords.

While the data may not be strictly representative of the population of all types of hostile content, we do find it to reflect a broad array of hostile content among a demographic that is highly vulnerable to the effects of online hostility.

\section{Conclusion}
\label{sec.conclusion}
We proposed methods to forecast both the presence and intensity of hostility in Instagram comments. Using a combination of linguistic and social features, the best model produces an AUC of 0.82 for forecasting the presence of hostility ten or more hours in the future, and an AUC of 0.91 for forecasting whether a post will receive more than 10 hostile comments or only one hostile comment. We find several predictors of future hostility, including (1) the post's author has received hostile comments in the past; (2) the use of user-directed profanity; (3) the number of distinct users participating in a conversation; and (4) trends in hostility thus far in the conversation.

By distinguishing between posts that will receive many hostile comments and those that will receive few or none, the methods proposed here provide new ways to prioritize specific posts for intervention. Since moderation resources are limited, it makes sense to assign them to posts where there is still time to de-escalate a conversation or prevent additional commenting. For instance, Instagram and similar platforms could use our approach to manage their moderation queues. Similarly, Instagram may use features our approach identifies (e.g., prior hostilities, number of distinct users) to improve their existing comment management features---offensive comment filtering and comment controls.

There are several avenues for future work. Given that Instagram is primarily a photo-sharing site, it is possible that image classification algorithms can be used to identify image attributes that predict hostile comments. Second, given the predictive power of the previous post feature, inferring more detailed user attributes (e.g., age, gender, ethnicity, etc.) may provide additional context for forecasting. Third, it may be possible to extract from these data insights into the effectiveness of de-escalation strategies --- e.g., given a hostile comment, are there certain responses that diffuse the situation, as opposed to leading to increased hostilities? Argumentation mining methods may be applicable here~\cite{palau2009argumentation}. Finally, we are experimenting with user-facing tools that allow users to monitor their own feeds (or their children's) for brewing escalations, enabling them to take action on- or offline.

\section{Acknowledgements}
This material is based upon work supported by the Nayar Prize at the Illinois Institute of Technology and the National Science Foundation under Grant No. 1822228.

\bibliography{ref.bib}

\begin{thebibliography}{}

\bibitem[\protect\citeauthoryear{Al-garadi, Varathan, and
  Ravana}{2016}]{al2016cybercrime}
Al-garadi, M.~A.; Varathan, K.~D.; and Ravana, S.~D.
\newblock 2016.
\newblock Cybercrime detection in online communications: The experimental case
  of cyberbullying detection in the twitter network.
\newblock {\em Computers in Human Behavior} 63:433--443.

\bibitem[\protect\citeauthoryear{Bellmore \bgroup et al\mbox.\egroup
  }{2015}]{bellmore2015five}
Bellmore, A.; Calvin, A.~J.; Xu, J.-M.; and Zhu, X.
\newblock 2015.
\newblock The five w’s of “bullying” on twitter: who, what, why, where,
  and when.
\newblock {\em Computers in human behavior} 44:305--314.

\bibitem[\protect\citeauthoryear{Bojanowski \bgroup et al\mbox.\egroup
  }{2016}]{bojanowski2016enriching}
Bojanowski, P.; Grave, E.; Joulin, A.; and Mikolov, T.
\newblock 2016.
\newblock Enriching word vectors with subword information.
\newblock {\em arXiv preprint arXiv:1607.04606}.

\bibitem[\protect\citeauthoryear{Chen \bgroup et al\mbox.\egroup
  }{2012}]{Chen2012-qc}
Chen, Y.; Zhou, Y.; Zhu, S.; and Xu, H.
\newblock 2012.
\newblock Detecting offensive language in social media to protect adolescent
  online safety.
\newblock In {\em Privacy, Security, Risk and Trust ({PASSAT)}, 2012
  International Conference on and 2012 International Conference on Social
  Computing ({SocialCom})},  71--80.
\newblock IEEE.

\bibitem[\protect\citeauthoryear{Cheng \bgroup et al\mbox.\egroup
  }{2017}]{Cheng2017Anyone}
Cheng, J.; Bernstein, M.; Danescu-Niculescu-Mizil, C.; and Leskovec, J.
\newblock 2017.
\newblock Anyone can become a troll: Causes of trolling behavior in online
  discussions.
\newblock  1217--1230.
\newblock ACM Press.

\bibitem[\protect\citeauthoryear{Crawford and
  Gillespie}{2014}]{Crawford2014What}
Crawford, K., and Gillespie, T.
\newblock 2014.
\newblock What is a flag for? social media reporting tools and the vocabulary
  of complaint.
\newblock {\em New Media \& Society}  1461444814543163.

\bibitem[\protect\citeauthoryear{Dadvar \bgroup et al\mbox.\egroup
  }{2012}]{dadvar2012improved}
Dadvar, M.; de~Jong, F.; Ordelman, R.; and Trieschnigg, R.
\newblock 2012.
\newblock Improved cyberbullying detection using gender information.
\newblock In {\em Proceedings of the Twelfth Dutch-Belgian Information
  Retrieval Workshop (DIR 2012)},  23--26.
\newblock University of Ghent.

\bibitem[\protect\citeauthoryear{Davidson \bgroup et al\mbox.\egroup
  }{2017}]{Davidson2017-xm}
Davidson, T.; Warmsley, D.; Macy, M.; and Weber, I.
\newblock 2017.
\newblock Automated hate speech detection and the problem of offensive
  language.

\bibitem[\protect\citeauthoryear{Dinakar, Reichart, and
  Lieberman}{2011}]{Dinakar2011-jr}
Dinakar, K.; Reichart, R.; and Lieberman, H.
\newblock 2011.
\newblock Modeling the detection of textual cyberbullying.
\newblock In {\em Fifth International {AAAI} Conference on Weblogs and
  SocialMedia"}.

\bibitem[\protect\citeauthoryear{Dovidio \bgroup et al\mbox.\egroup
  }{2013}]{Dovidio2013Sage}
Dovidio, J.~F.; Hewstone, M.; Glick, P.; and Esses, V.~M., eds.
\newblock 2013.
\newblock {\em The SAGE Handbook of Prejudice, Stereotyping and
  Discrimination}.
\newblock Los Angeles: SAGE Publications Ltd, 1 edition edition.

\bibitem[\protect\citeauthoryear{Duggan \bgroup et al\mbox.\egroup
  }{2014}]{duggan2014online}
Duggan, M.; Rainie, L.; Smith, A.; Funk, C.; Lenhart, A.; and Madden, M.
\newblock 2014.
\newblock Online harassment. washington, dc: Pew research center.

\bibitem[\protect\citeauthoryear{Duggan}{2017}]{Duggan2017ei}
Duggan, M.
\newblock 2017.
\newblock Online harassment 2017.
\newblock http://www.pewinternet.org/2017/07/11/online-harassment-2017/.
\newblock Accessed: 2018-1-26.

\bibitem[\protect\citeauthoryear{Geiger}{2016}]{Geiger2016-pi}
Geiger, R.~S.
\newblock 2016.
\newblock Bot-based collective blocklists in twitter: the counterpublic
  moderation of harassment in a networked public space.
\newblock {\em Information, Communication, and Society} 19(6):787--803.

\bibitem[\protect\citeauthoryear{Guberman and Hemphill}{2017}]{Guberman2017-sz}
Guberman, J., and Hemphill, L.
\newblock 2017.
\newblock Challenges in modifying existing scales for detecting harassment in
  individual tweets.
\newblock In {\em 50th Annual Hawaii International Conference on System
  Sciences ({HICSS-50})}.

\bibitem[\protect\citeauthoryear{Krebs}{1975}]{Krebs1975-yh}
Krebs, D.
\newblock 1975.
\newblock Empathy and altruism.
\newblock {\em J. Pers. Soc. Psychol.} 32(6):1134--1146.

\bibitem[\protect\citeauthoryear{M{\"a}rtens \bgroup et al\mbox.\egroup
  }{2015}]{Martens2015-bl}
M{\"a}rtens, M.; Shen, S.; Iosup, A.; and Kuipers, F.
\newblock 2015.
\newblock Toxicity detection in multiplayer online games.
\newblock In {\em 2015 International Workshop on Network and Systems Support
  for Games ({NetGames})},  1--6.

\bibitem[\protect\citeauthoryear{Mikolov \bgroup et al\mbox.\egroup
  }{2013}]{mikolov2013distributed}
Mikolov, T.; Sutskever, I.; Chen, K.; Corrado, G.~S.; and Dean, J.
\newblock 2013.
\newblock Distributed representations of words and phrases and their
  compositionality.
\newblock In {\em Advances in neural information processing systems},
  3111--3119.

\bibitem[\protect\citeauthoryear{Nakamura}{2015}]{Nakamura2015-mg}
Nakamura, L.
\newblock 2015.
\newblock The unwanted labour of social media: Women of colour call out culture
  as venture community management.
\newblock {\em New Formations} 86(86):106--112.

\bibitem[\protect\citeauthoryear{Nocentini \bgroup et al\mbox.\egroup
  }{2010}]{Nocentini2010-wt}
Nocentini, A.; Calmaestra, J.; Schultze-Krumbholz, A.; Scheithauer, H.; Ortega,
  R.; and Menesini, E.
\newblock 2010.
\newblock Cyberbullying: Labels, behaviours and definition in three european
  countries.
\newblock {\em Journal of Psychologists and Counsellors in Schools}
  20(2):129--142.

\bibitem[\protect\citeauthoryear{NORC}{2017}]{Sood2012new}
NORC.
\newblock 2017.
\newblock New survey: Snapchat and instagram are most popular social media
  platforms among american teens: Black teens are the most active on social
  media and messaging apps.
\newblock {\em ScienceDaily}.

\bibitem[\protect\citeauthoryear{Palau and
  Moens}{2009}]{palau2009argumentation}
Palau, R.~M., and Moens, M.-F.
\newblock 2009.
\newblock Argumentation mining: the detection, classification and structure of
  arguments in text.
\newblock In {\em Proceedings of the 12th international conference on
  artificial intelligence and law},  98--107.
\newblock ACM.

\bibitem[\protect\citeauthoryear{Pater \bgroup et al\mbox.\egroup
  }{2016}]{Pater2016-bv}
Pater, J.~A.; Kim, M.~K.; Mynatt, E.~D.; and Fiesler, C.
\newblock 2016.
\newblock Characterizations of online harassment: Comparing policies across
  social media platforms.
\newblock In {\em {GROUP}},  369--374.
\newblock jesspater.com.

\bibitem[\protect\citeauthoryear{Pfeil and Zaphiris}{2007}]{Pfeil2007-kd}
Pfeil, U., and Zaphiris, P.
\newblock 2007.
\newblock Patterns of empathy in online communication.
\newblock In {\em Proceedings of the {SIGCHI} conference on Human factors in
  computing systems - {CHI} '07},  919.
\newblock New York, New York, USA: ACM Press.

\bibitem[\protect\citeauthoryear{Phillips and Milner}{2017}]{Phillips2017-ep}
Phillips, W., and Milner, R.~M.
\newblock 2017.
\newblock We are all internet bullies.
\newblock {\em Time}.

\bibitem[\protect\citeauthoryear{Phillips}{2015}]{Phillips2015-lr}
Phillips, W.
\newblock 2015.
\newblock {\em This Is Why We Can't Have Nice Things: Mapping the Relationship
  between Online Trolling and Mainstream Culture}.
\newblock The MIT Press.

\bibitem[\protect\citeauthoryear{Preece}{1998}]{Preece1998-xk}
Preece, J.
\newblock 1998.
\newblock Empathic communities: Reaching out across the web.
\newblock {\em Interactions} 5(2):32--43.

\bibitem[\protect\citeauthoryear{Reynolds, Kontostathis, and
  Edwards}{2011}]{Reynolds2011-xt}
Reynolds, K.; Kontostathis, A.; and Edwards, L.
\newblock 2011.
\newblock Using machine learning to detect cyberbullying.
\newblock In {\em 10th International Conference on Machine Learning and
  Applications and Workshops ({ICMLA})}, volume~2,  241--244.
\newblock IEEE.

\bibitem[\protect\citeauthoryear{Roberts}{2016}]{Roberts2016Commercial}
Roberts, S.
\newblock 2016.
\newblock Commercial content moderation: Digital laborers' dirty work.
\newblock In Noble, S.~U., and Tynes, B.~M., eds., {\em The Intersectional
  Internet: Race, Sex, Class and Culture Online}. New York: Peter Lang
  Publishing.

\bibitem[\protect\citeauthoryear{Rubin, Pruitt, and Kim}{1994}]{Rubin1994-mi}
Rubin, J.~Z.; Pruitt, D.~G.; and Kim, S.~H.
\newblock 1994.
\newblock {\em Social conflict: Escalation, stalemate, and settlement}.
\newblock New York: McGraw Hill Book Company, 2nd edition.

\bibitem[\protect\citeauthoryear{Sood, Antin, and
  Churchill}{2012}]{Sood2012-cn}
Sood, S.; Antin, J.; and Churchill, E.
\newblock 2012.
\newblock Profanity use in online communities.
\newblock In {\em Proceedings of the {SIGCHI} Conference on Human Factors in
  Computing Systems}, CHI '12,  1481--1490.
\newblock New York, NY, USA: ACM.

\bibitem[\protect\citeauthoryear{Sood, Churchill, and
  Antin}{2012}]{Sood2012-jt}
Sood, S.~O.; Churchill, E.~F.; and Antin, J.
\newblock 2012.
\newblock Automatic identification of personal insults on social news sites.
\newblock {\em J. Am. Soc. Inf. Sci. Technol.} 63(2):270--285.

\bibitem[\protect\citeauthoryear{Su and Khoshgoftaar}{2009}]{Su2009-mx}
Su, X., and Khoshgoftaar, T.~M.
\newblock 2009.
\newblock A survey of collaborative filtering techniques.
\newblock {\em Advances in Artificial Intelligence} 2009:4:2--4:2.

\bibitem[\protect\citeauthoryear{Thompson}{2017}]{Thompson2017-gx}
Thompson, N.
\newblock 2017.
\newblock The great tech panic: Instagram's kevin systrom wants to clean up the
  internet.
\newblock {\em Wired}.

\bibitem[\protect\citeauthoryear{Wang and Cardie}{2016}]{Wang2016Piece}
Wang, L., and Cardie, C.
\newblock 2016.
\newblock A piece of my mind: A sentiment analysis approach for online dispute
  detection.
\newblock {\em arXiv preprint arXiv:1606.05704}.

\bibitem[\protect\citeauthoryear{Wang \bgroup et al\mbox.\egroup
  }{2014}]{Wang2014-ir}
Wang, W.; Chen, L.; Thirunarayan, K.; and Sheth, A.~P.
\newblock 2014.
\newblock Cursing in english on twitter.
\newblock In {\em Proceedings of the 17th {ACM} conference on Computer
  supported cooperative work \& social computing},  415--425.
\newblock ACM.

\bibitem[\protect\citeauthoryear{Yang and Leskovec}{2011}]{yang2011patterns}
Yang, J., and Leskovec, J.
\newblock 2011.
\newblock Patterns of temporal variation in online media.
\newblock In {\em Proceedings of the fourth ACM international conference on Web
  search and data mining},  177--186.
\newblock ACM.

\bibitem[\protect\citeauthoryear{Yin \bgroup et al\mbox.\egroup
  }{2009}]{Yin2009Detection}
Yin, D.; Xue, Z.; Hong, L.; Davison, B.~D.; Kontostathis, A.; and Edwards, L.
\newblock 2009.
\newblock Detection of harassment on web 2.0.
\newblock {\em Proceedings of the Content Analysis in the WEB} 2:1--7.

\end{thebibliography}
\bibliographystyle{aaai}

\end{document}